\documentclass[11pt,letterpaper]{article}
\usepackage{ijcnlp2017}
\usepackage{times}
\usepackage{latexsym}
\usepackage{paralist,relsize}

\ijcnlpfinalcopy

\def\ijcnlppaperid{156}

\usepackage{xspace}
\usepackage{mathtools}
\usepackage{amsmath}
\usepackage{amssymb}
\usepackage{color}
\usepackage{multirow}
\usepackage{tikz}
\usepackage{pifont}
\usepackage[inline]{enumitem}
\usetikzlibrary{positioning}
\usetikzlibrary{calc}
\usetikzlibrary{fit}
\usetikzlibrary{decorations.text}
\usetikzlibrary{shapes.misc}
\usetikzlibrary{shapes.geometric,arrows}
\usepackage{colortbl}
\usepackage{mathrsfs}
\usepackage{bm}
\usepackage{booktabs}
\usepackage{pgfplots}
\usepackage{subcaption}
\usepackage{arydshln}

\newcommand{\eg}{e.g.\ }
\newcommand{\ie}{i.e.\ }

\newcommand{\system}[2][]{\textsc{#2}#1\xspace}
\newcommand{\class}[2][]{\texttt{#2}#1\xspace}

\newcommand{\secref}[1]{Section~\ref{#1}\xspace}
\newcommand{\tabref}[2][]{Table#1~\ref{#2}\xspace}
\newcommand{\figref}[2][]{Figure#1~\ref{#2}\xspace}

\newcommand{\equref}[2][]{Equation#1~\bracketref{#2}}
\newcommand{\bracketref}[1]{(\ref{#1})\xspace}

\newcommand{\bb}[1]{\mathbb{#1}}
\newcommand{\R}{\bb{R}}
\newcommand{\mat}[2][]{\boldsymbol{#2}_{#1}}
\newcommand{\matfwd}[2][]{\overrightarrow{\bm{#2}}_{#1}}
\newcommand{\matrev}[2][]{\overleftarrow{\bm{#2}}_{#1}}

\renewcommand{\vec}[2][]{\boldsymbol{#2}^{#1}}
\newcommand{\vecfwd}[2][]{\overrightarrow{\bm{#2}}^{#1}}
\newcommand{\vecrev}[2][]{\overleftarrow{\bm{#2}}^{#1}}

\newcommand{\gru}{\textrm{GRU}}
\newcommand{\grufwd}{\overrightarrow{\textrm{GRU}}}
\newcommand{\grurev}{\overleftarrow{\textrm{GRU}}}
\newcommand{\softmax}{\textrm{softmax}}
\newcommand{\xentropy}{\ensuremath{\operatorname{CrossEntropy}}\xspace}

\newcommand{\T}{\mathstrut\scriptscriptstyle\top}
\newcommand{\tran}{^{\T}}

\newcommand{\memnn}[1][]{\system[#1]{MemNet}}

\newcommand{\crf}[1][]{\system[#1]{CRF}}
\newcommand{\mecrf}[1][]{\system[#1]{ME-CRF}}
\newcommand{\lstm}[1][]{\system[#1]{LSTM}}
\newcommand{\bilstm}[1][]{\system[#1]{Bi-LSTM}}
\newcommand{\rnn}[1][]{\system[#1]{RNN}}
\newcommand{\convcrf}[1][]{\system[#1]{Conv-CRF}}
\newcommand{\lstmcrf}[1][]{\system[#1]{LSTM-CRF}}
\newcommand{\bilstmcrf}[1][]{\system[#1]{Bi-LSTM-CRF}}

\newcommand{\kim}{\system{CRF$_{\text{Kim}}$}}
\newcommand{\wangcrf}{\system{CRF$_{\text{Wang}}$}}
\newcommand{\wangparser}{\system{DepParser}}

\newcommand{\RNum}[1]{\uppercase\expandafter{\romannumeral #1\relax}}

\newcommand{\ex}[1]{\textit{#1}\xspace}

\DeclareMathOperator*{\argmax}{arg\,max}

\title{Capturing Long-range Contextual Dependencies with\\ Memory-enhanced Conditional Random Fields}

 \author{Fei Liu \qquad Timothy Baldwin \qquad Trevor Cohn \\
         School of Computing and Information Systems \\ The University of Melbourne \\ Victoria, Australia\\
         {\tt {fliu3@student.unimelb.edu.au}} \\
         {\tt {tb@ldwin.net}\,\,\, {t.cohn@unimelb.edu.au}}}

\date{}

\begin{document}

\maketitle

\begin{abstract}
  Despite successful applications across a broad range of NLP tasks, conditional random fields (``\crf[s]''), in particular the linear-chain variant, are only able to model local features.
  While this has important benefits in terms of inference tractability, it limits the ability of the model to capture long-range dependencies between items.
  Attempts to extend \crf[s] to capture long-range dependencies have largely come at the cost of computational complexity and approximate inference.
  In this work, we propose an extension to \crf[s] by integrating external memory, taking inspiration from memory networks, thereby allowing \crf[s] to incorporate information far beyond neighbouring steps.
  Experiments across two tasks show substantial improvements over strong \crf and \lstm baselines.
\end{abstract}

\section{Introduction}
\label{sec:intro}

\begin{figure*}[t]
\begin{center}
\resizebox{\textwidth}{!}{
\begin{tikzpicture}

%
%
%
%
%
%

\node[ellipse, draw, fill=black!20, text height =3mm, minimum height=1cm,minimum width = 2cm,align=center] at (0.0, 0.0)  (x0) {Interfax};
\node[ellipse, draw, fill=white, text height =3mm, minimum height=1cm, minimum width = 2cm,align=center] [above=0.5 of x0]  (y0) {B-ORG};
\draw (x0.north) -- (y0.south);
\node[text height =3mm, minimum height=1cm, minimum width =.6cm,align=center] [left=0.3 of y0]  (y_1) {$\cdots$};
\draw (y_1.east) -- (y0.west);

\node[ellipse, draw, fill=black!20, text height =3mm, minimum height=1cm, minimum width = 2cm,align=center] [right=0.3 of x0] (x1) {news};
\node[ellipse, draw, fill=white, text height =3mm, minimum height=1cm, minimum width = 2cm,align=center] [above=0.5 of x1]  (y1) {O};
\draw (x1.north) -- (y1.south);
\draw (y0.east) -- (y1.west);

\node[ellipse, draw, fill=black!20, text height =3mm, minimum height=1cm, minimum width = 2cm,align=center] [right=0.3 of x1] (x2) {agency};
\node[ellipse, draw, fill=white, text height =3mm, minimum height=1cm, minimum width = 2cm,align=center] [above=0.5 of x2]  (y2) {O};
\draw (x2.north) -- (y2.south);
\draw (y1.east) -- (y2.west);

\node[ellipse, draw, fill=black!20, text height =3mm, minimum height=1cm, minimum width = 2cm,align=center] [right=0.3 of x2] (x3) {said};
\node[ellipse, draw, fill=white, text height =3mm, minimum height=1cm, minimum width = 2cm,align=center] [above=0.5 of x3]  (y3) {O};
\draw (x3.north) -- (y3.south);
\draw (y2.east) -- (y3.west);

\node[text height =3mm, minimum height=1cm, minimum width=.6cm, align=center] [right=0.5 of y3]  (y4) {$\cdots$};
\draw (y3.east) -- (y4.west);
\draw [dashed] ($(y4.north)+(0.0,0.2)$) -- ($(y4.south)+(0.0,-1.4)$);

\node[ellipse, draw, fill=black!20, text height =3mm, minimum height=1cm, minimum width = 2cm,align=center] [right=1.6 of x3] (x5) {Interfax};
\node[ellipse, draw, fill=white, text height =3mm, minimum height=1cm, minimum width = 2cm,align=center] [above=0.5 of x5]  (y5) {B-ORG};
\draw (x5.north) -- (y5.south);
\draw (y4.east) -- (y5.west);

\node[ellipse, draw, fill=black!20, text height =3mm, minimum height=1cm, minimum width = 2cm,align=center] [right=0.3 of x5] (x6) {quoted};
\node[ellipse, draw, fill=white, text height =3mm, minimum height=1cm, minimum width = 2cm,align=center] [above=0.5 of x6]  (y6) {O};
\draw (x6.north) -- (y6.south);
\draw (y5.east) -- (y6.west);

\node[ellipse, draw, fill=black!20, text height =3mm, minimum height=1cm, minimum width = 2cm,align=center] [right=0.3 of x6] (x7) {Russian};
\node[ellipse, draw, fill=white, text height =3mm, minimum height=1cm, minimum width = 2cm,align=center] [above=0.5 of x7]  (y7) {B-MISC};
\draw (x7.north) -- (y7.south);
\draw (y6.east) -- (y7.west);

\node[ellipse, draw, fill=black!20, text height =3mm, minimum height=1cm, minimum width = 2cm,align=center] [right=0.3 of x7] (x8) {military};
\node[ellipse, draw, fill=white, text height =3mm, minimum height=1cm, minimum width = 2cm,align=center] [above=0.5 of x8]  (y8) {O};
\draw (x8.north) -- (y8.south);
\draw (y7.east) -- (y8.west);

\node[text height =3mm, minimum height=1cm, minimum width =.6cm,align=center] [right=0.3 of y8]  (y9) {$\cdots$};
\draw (y8.east) -- (y9.west);

\end{tikzpicture}
}
\end{center}
\caption{A NER example with long-range contextual dependencies. The vertical dash line indicates a sentence boundary.}
\label{fig:nerexample}
\end{figure*}
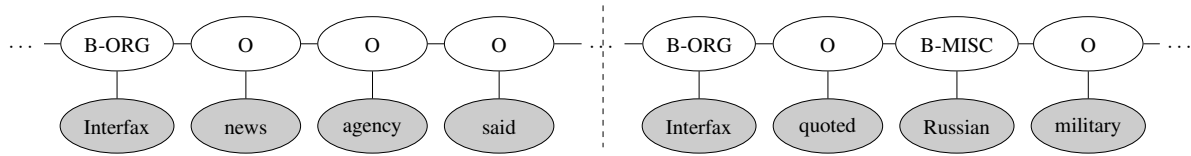

While long-range contextual dependencies are prevalent in natural language, 
for tractability reasons, most statistical models capture only local features \cite{Finkel+:2005}. Take the sentence in \figref{fig:nerexample}, for example. Here, while it is easy to determine that \ex{Interfax} in the second sentence is a named entity, it is hard to determine its semantic class, as there is little context information. The usage in the first sentence, on the other hand, can be reliably disambiguated due to the post-modifying phase \ex{news agency}. Ideally we would like to be able to share such contexts across all usages (and variants) of a given named entity for reliable and consistent identification and disambiguation. 

A related example is forum thread discourse analysis. Previous work has largely focused on linear-chain Conditional Random Fields (\crf[s]) \cite{Wang+:2011,Zhang+:2017}, framing the task as one of sequence tagging. Although \crf[s] are adept at capturing local structure, the problem does not naturally suit a linear sequential structure, \ie, a post may be a reply to either a neighbouring post or one posted far earlier within the same thread. In both cases, contextual dependencies can be long-range, necessitating the ability to capture dependencies between arbitrarily distant items. Identifying this key limitation, \newcite{Sutton+:2004} and \newcite{Finkel+:2005} proposed the use of \crf[s] with skip connections to incorporate long-range dependencies. In both cases the graph structure must be supplied a priori, rather than learned, and both techniques incur the need for costly approximate inference.

Recurrent neural networks (\rnn[s]) have been proposed as an alternative technique for encoding sequential inputs, however plain \rnn[s] are unable to  capture long-range dependencies \cite{Bengio+:1994,Hochreiter+:2001} and variants such as \lstm[s] \cite{Hochreiter+:1997}, although more capabable of capturing non-local patterns, still exhibit a significant locality bias in practice \cite{Lai+:2015,Linzhen+:2016}. 

In this paper, taking inspiration from the work of \newcite{Weston+:2015} on memory networks (\memnn[s]), we propose to extend \crf[s] by integrating external memory mechanisms, thereby enabling the model to look beyond localised features and have access to the entire sequence. This is achieved with attention over every entry in the memory. Experiments on named entity recognition and forum thread parsing, both of which involve long-range contextual dependencies, demonstrate the effectiveness of the proposed model, achieving state-of-the-art performance on the former, and outperforming a number of strong baselines in the case of the latter. 
A full implementation of the model is available at: \url{https://github.com/liufly/mecrf}.

The paper is organised as follows: after reviewing previous studies on capturing long range contextual dependencies and related models in \secref{sec:bg}, we detail the elements of the proposed model in \secref{sec:model}. \secref{sec:expforum} and \ref{sec:expner} present the experimental results on two different datasets: one for thread discourse structure prediction and the other named entity recognition (NER), with analyses and visualisation in their respective sections. Lastly, \secref{sec:conc} concludes the paper.


\section{Related Work}
\label{sec:bg}

In this section, we review the different families of models that are
relevant to this work, in capturing long-range contextual dependencies
in different ways.

\paragraph{Conditional Random Fields (\crf[s]).}
\crf[s] \cite{Lafferty+:2001}, in particular linear-chain CRFs, have been widely adopted and applied to sequence labelling tasks in NLP, but have the critical limitation that they only capture local structure \cite{Sutton+:2004,Finkel+:2005}, despite non-local structure being common in structured language classification tasks. In the context of named entity recognition (``NER''), \newcite{Sutton+:2004} proposed skip-chain \crf[s] as a means of alleviating this shortcoming, wherein distant items are connected in a sequence based on a heuristic such as string identity (to achieve label consistency across all instances of the same string).
The idea of label consistency and exploiting non-local features has also been explored in the work of \newcite{Finkel+:2005}, who take long-range structure into account while maintaining tractable inference with Gibbs sampling \cite{Geman+:1984}, by performing approximate inference over factored probabilistic models. While both of these lines of work report impressive results on information extraction tasks, they come at the price of high computational cost and incompatibility with exact inference.

Similar ideas have also been explored by \newcite{Krishnan+:2006} for NER, where they apply two \crf[s], the first of which makes predictions based on local information, and the second combines named entities identified by the first CRF in a single cluster, thereby enforcing label consistency and enabling the use of a richer set of features to capture non-local dependencies. \newcite{Liao+:2010} make a strong case for going beyond sentence boundaries and leveraging document-level information for event extraction. 

While we take inspiration from these earlier studies, we do not enforce label consistency as a hard constraint, and additionally do not sacrifice inference tractability: our model is capable of incorporating non-local features, and is compatible with exact inference methods.

\paragraph{Recurrent Neural Networks (\rnn[s]).}
Recently, the broad adoption of deep learning methods in NLP has given rise to the prevalent use of \rnn[s]. Long short-term memories (``\lstm[s]'': \newcite{Hochreiter+:1997}), a particular variant of \rnn, have become particularly popular, and been successfully applied to a large number of tasks: speech recognition \cite{Graves+:2013}, sequence tagging \cite{Huang+:2015}, document categorisation \cite{Yang+:2016}, and  machine translation \cite{Cho+:2014b,Bahdanau+:2014}. However, as pointed out by \newcite{Lai+:2015} and \newcite{Linzhen+:2016}, \rnn[s] --- including \lstm[s] --- are biased towards immediately preceding (or neighbouring, in the case of bi-directional \rnn[s]) items, and perform poorly in contexts which involve long-range contextual dependencies, despite the inclusion of memory cells. This is further evidenced by the work of \newcite{Cho+:2014b}, who show that the performance of a basic encoder--decoder deteriorates as the length of the input sentence increases.

\paragraph{Memory networks (\memnn[s]).}
More recently, \newcite{Weston+:2015} proposed memory networks and showed that the augmentation of memory is crucial to performing inference requiring long-range dependencies, especially when document-level reasoning between multiple supporting facts is required. Of particular interest to our work are so-called ``memory hops'' in memory networks, which are guided by an attention mechanism based on the relevance between a question and each supporting context sentence in the memory hop. Governed by the attention mechanism, the ability to access the entire sequence is similar to the soft alignment idea proposed by \newcite{Bahdanau+:2014} for neural machine translation. In this work, we borrow the concept of memory hops and integrate it into \crf[s], thereby enabling the model to look beyond localised features and have access to the whole sequence via an attention mechanism.


\section{Methodology}
\label{sec:model}

In the context of sequential tagging, we assume the input is in the form of sequence pairs: 
$\mathcal{D} = \{\mathbf{x}^{(n)}, \mathbf{y}^{(n)}\}_{n=1}^{N}$ 
where $\mathbf{x}^{(n)}$ is the input of the $n$-th example in dataset $\mathcal{D}$ and consists of a sequence: $\{x_{1}^{(n)}, x_2^{(n)}, \ldots, x_T^{(n)}\}$. Similarly, $\mathbf{y}^{(n)}$ is of the same length as $\mathbf{x}^{(n)}$ and consists of the corresponding labels $\{y_1^{(n)}, y_2^{(n)}, \ldots, y_T^{(n)}\}$. 
For notational convenience, hereinafter we omit the superscript denoting the $n$-th example.

In the case of NER, each $x_t$ is a word in a sentence with $y_t$ being the corresponding NER label. For forum thread discourse analysis, $x_t$ represents the text of an entire post whereas $y_t$ is the dialogue act label for the $t$-th post.

The proposed model extends \crf[s] by integrating external memory and is therefore named a \textbf{Memory-Enhanced Conditional Random Field} (``\mecrf''). We take inspiration from Memory Networks (``\memnn[s]'': \newcite{Weston+:2015}) and incorporate so-called memory hops into \crf[s], thereby allowing the model to have unrestricted access to the whole sequence rather than localised features as in RNNs \cite{Lai+:2015,Linzhen+:2016}. 

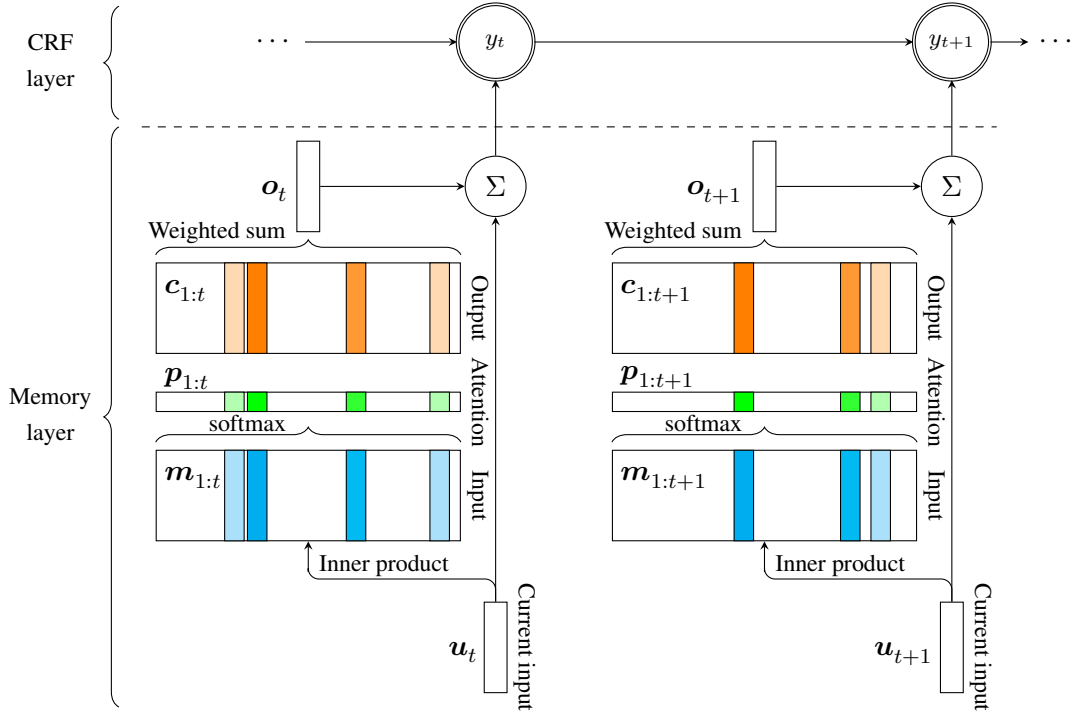
\begin{figure*}[tb]
\begin{center}
\begin{tikzpicture}

\node[draw,align=center,minimum height=1.2cm,minimum width=0.3cm] at (6.0,0.0) (u_t) {};
\node[anchor=north west,shift={(-6mm,2mm)}] (u_t_label) at (u_t.west){$\vec{u}_{t}$};

\node[draw,align=center,minimum height=1.2cm,minimum width=4.0cm]  [above left=0.8cm and 0.3cm of u_t] (m_i) {};
\node[draw,align=center,minimum height=0.2cm,minimum width=4.0cm]  [above=0.5cm of m_i] (p_i) {};
\node[draw,align=center,minimum height=1.2cm,minimum width=4.0cm]  [above=0.5cm of p_i] (c_i) {};
\node[anchor=north west,shift={(-0.2mm,-1mm)}] (m_i_label) at (m_i.north west){$\vec{m}_{1:t}$};
\node[anchor=south west,shift={(-0.2mm,-1mm)}] (p_i_label) at (p_i.north west){$\vec{p}_{1:t}$};
\node[anchor=north west,shift={(-0.2mm,-1mm)}] (c_i_label) at (c_i.north west){$\vec{c}_{1:t}$};
\node[anchor=north,shift={(5mm,0mm)},rotate=-90] (m_i_label) at (m_i.east){\footnotesize Input};
\node[anchor=north,shift={(5mm,0mm)},rotate=-90] (p_i_label) at (p_i.east){\footnotesize Attention};
\node[anchor=north,shift={(5mm,0mm)},rotate=-90] (c_i_label) at (c_i.east){\footnotesize Output};
\node[anchor=north,shift={(5mm,0mm)},rotate=-90] (u_t_label) at (u_t.east){\footnotesize Current input};

\node[draw,anchor=north west,shift={(9mm,0mm)},minimum height=1.2cm,minimum width=0.05mm,fill=cyan!30] (m_1) at (m_i.north west) {};
\node[draw,anchor=north west,shift={(12mm,0mm)},minimum height=1.2cm,minimum width=0.05mm,fill=cyan!100] (m_2) at (m_i.north west) {};
\node[draw,anchor=north west,shift={(25mm,0mm)},minimum height=1.2cm,minimum width=0.05mm,fill=cyan!80] (m_3) at (m_i.north west) {};
\node[draw,anchor=north west,shift={(36mm,0mm)},minimum height=1.2cm,minimum width=0.05mm,fill=cyan!30] (m_4) at (m_i.north west) {};
\node[draw,anchor=north west,shift={(9mm,0mm)},minimum height=0.2cm,minimum width=0.05mm,fill=green!30] (p_1) at (p_i.north west) {};
\node[draw,anchor=north west,shift={(12mm,0mm)},minimum height=0.2cm,minimum width=0.05mm,fill=green!100] (p_2) at (p_i.north west) {};
\node[draw,anchor=north west,shift={(25mm,0mm)},minimum height=0.2cm,minimum width=0.05mm,fill=green!80] (p_3) at (p_i.north west) {};
\node[draw,anchor=north west,shift={(36mm,0mm)},minimum height=0.2cm,minimum width=0.05mm,fill=green!30] (p_4) at (p_i.north west) {};
\node[draw,anchor=north west,shift={(9mm,0mm)},minimum height=1.2cm,minimum width=0.05mm,fill=orange!30] (c_1) at (c_i.north west) {};
\node[draw,anchor=north west,shift={(12mm,0mm)},minimum height=1.2cm,minimum width=0.05mm,fill=orange!100] (c_2) at (c_i.north west) {};
\node[draw,anchor=north west,shift={(25mm,0mm)},minimum height=1.2cm,minimum width=0.05mm,fill=orange!80] (c_3) at (c_i.north west) {};
\node[draw,anchor=north west,shift={(36mm,0mm)},minimum height=1.2cm,minimum width=0.05mm,fill=orange!30] (c_4) at (c_i.north west) {};

\draw[decorate,decoration={brace,amplitude=2mm,raise=1mm},yshift=0pt](m_i.north west) -- (m_i.north east) node [black,midway,xshift=-0.8cm,yshift=0.35cm] {\footnotesize $\softmax$};
\draw[decorate,decoration={brace,amplitude=2mm,raise=1mm},yshift=0pt](c_i.north west) -- (c_i.north east) node [black,midway,xshift=-1.2cm,yshift=0.4cm] {\footnotesize Weighted sum};

\draw [rounded corners,->,>=stealth] (u_t.north) |- ($(m_i.south) + (0.0,-0.5)$) -| node [midway,above,shift={(10mm,-1mm)}] (u2m_label) {\footnotesize Inner product} (m_i.south);

\node[draw,align=center,minimum height=1.2cm,minimum width=0.3cm] [above=0.4cm of c_i] (o_t) {};
\node[anchor=north west,shift={(-6mm,2mm)}] (u_t_label) at (o_t.west){$\vec{o}_{t}$};

\node[draw,align=center,circle,minimum size=0.8cm] [above=5.1cm of u_t] (Sum1) {$\Sigma$};

\draw [rounded corners,->,>=stealth] (o_t.east) -- (Sum1.west);
\draw [rounded corners,->,>=stealth] (u_t.north) -- (Sum1.south);


\node[draw,align=center,minimum height=1.2cm,minimum width=0.3cm] at (12.0,0.0) (u_t1) {};
\node[anchor=north west,shift={(-10mm,2mm)}] (u_t1_label) at (u_t1.west){$\vec{u}_{t+1}$};

\node[draw,align=center,minimum height=1.2cm,minimum width=4.0cm]  [above left=0.8cm and 0.3cm of u_t1] (m_i1) {};
\node[draw,align=center,minimum height=0.2cm,minimum width=4.0cm]  [above=0.5cm of m_i1] (p_i1) {};
\node[draw,align=center,minimum height=1.2cm,minimum width=4.0cm]  [above=0.5cm of p_i1] (c_i1) {};
\node[anchor=north west,shift={(-0.2mm,-1mm)}] (m_i1_label) at (m_i1.north west){$\vec{m}_{1:t+1}$};
\node[anchor=south west,shift={(-0.2mm,-1mm)}] (p_i1_label) at (p_i1.north west){$\vec{p}_{1:t+1}$};
\node[anchor=north west,shift={(-0.2mm,-1mm)}] (c_i1_label) at (c_i1.north west){$\vec{c}_{1:t+1}$};
\node[anchor=north,shift={(5mm,0mm)},rotate=-90] (m_i1_label) at (m_i1.east){\footnotesize Input};
\node[anchor=north,shift={(5mm,0mm)},rotate=-90] (p_i1_label) at (p_i1.east){\footnotesize Attention};
\node[anchor=north,shift={(5mm,0mm)},rotate=-90] (c_i1_label) at (c_i1.east){\footnotesize Output};
\node[anchor=north,shift={(5mm,0mm)},rotate=-90] (u_t1_label) at (u_t1.east){\footnotesize Current input};

\node[draw,anchor=north west,shift={(16mm,0mm)},minimum height=1.2cm,minimum width=0.05mm,fill=cyan!100] (m_2) at (m_i1.north west) {};
\node[draw,anchor=north west,shift={(30mm,0mm)},minimum height=1.2cm,minimum width=0.05mm,fill=cyan!80] (m_3) at (m_i1.north west) {};
\node[draw,anchor=north west,shift={(34mm,0mm)},minimum height=1.2cm,minimum width=0.05mm,fill=cyan!30] (m_4) at (m_i1.north west) {};
\node[draw,anchor=north west,shift={(16mm,0mm)},minimum height=0.2cm,minimum width=0.05mm,fill=green!100] (p_2) at (p_i1.north west) {};
\node[draw,anchor=north west,shift={(30mm,0mm)},minimum height=0.2cm,minimum width=0.05mm,fill=green!80] (p_3) at (p_i1.north west) {};
\node[draw,anchor=north west,shift={(34mm,0mm)},minimum height=0.2cm,minimum width=0.05mm,fill=green!30] (p_4) at (p_i1.north west) {};
\node[draw,anchor=north west,shift={(16mm,0mm)},minimum height=1.2cm,minimum width=0.05mm,fill=orange!100] (c_2) at (c_i1.north west) {};
\node[draw,anchor=north west,shift={(30mm,0mm)},minimum height=1.2cm,minimum width=0.05mm,fill=orange!80] (c_3) at (c_i1.north west) {};
\node[draw,anchor=north west,shift={(34mm,0mm)},minimum height=1.2cm,minimum width=0.05mm,fill=orange!30] (c_4) at (c_i1.north west) {};

\draw[decorate,decoration={brace,amplitude=2mm,raise=1mm},yshift=0pt](m_i1.north west) -- (m_i1.north east) node [black,midway,xshift=-0.8cm,yshift=0.35cm] {\footnotesize $\softmax$};
\draw[decorate,decoration={brace,amplitude=2mm,raise=1mm},yshift=0pt](c_i1.north west) -- (c_i1.north east) node [black,midway,xshift=-1.2cm,yshift=0.4cm] {\footnotesize Weighted sum};

\draw [rounded corners,->,>=stealth] (u_t1.north) |- ($(m_i1.south) + (0.0,-0.5)$) -| node [midway,above,shift={(10mm,-1mm)}] (u2m1_label) {\footnotesize Inner product} (m_i1.south);

\node[draw,align=center,minimum height=1.2cm,minimum width=0.3cm] [above=0.4cm of c_i1] (o_t1) {};
\node[anchor=north west,shift={(-10mm,2mm)}] (u_t1_label) at (o_t1.west){$\vec{o}_{t+1}$};

\node[draw,align=center,circle,minimum size=0.8cm] [above=5.1cm of u_t1] (Sum2) {$\Sigma$};

\draw [rounded corners,->,>=stealth] (o_t1.east) -- (Sum2.west);
\draw [rounded corners,->,>=stealth] (u_t1.north) -- (Sum2.south);

\node[fit=(u_t) (o_t) (Sum1) (c_i) (u_t1) (Sum2),inner sep=5,minimum width=4.0cm] (memory_layer_frame) {};
\draw [dashed] (memory_layer_frame.north west) -- (memory_layer_frame.north east);


\draw[decorate,decoration={brace,amplitude=2mm,mirror,raise=3mm},yshift=0pt](memory_layer_frame.north west) -- (memory_layer_frame.south west) node [black,midway,xshift=-1.2cm,yshift=0cm,align=center] {\footnotesize Memory\\ \footnotesize layer};

\node[draw,double,align=center,circle,minimum size=1.0cm] [above=1.0cm of Sum1] (y_t) {\footnotesize $y_t$};
\node[draw,double,align=center,circle,minimum size=1.0cm] [above=1.0cm of Sum2] (y_t1) {\footnotesize $y_{t+1}$};
\draw [rounded corners,->,>=stealth] (Sum1.north) -- (y_t.south);
\draw [rounded corners,->,>=stealth] (Sum2.north) -- (y_t1.south);
\draw [rounded corners,->,>=stealth] (y_t.east) -- (y_t1.west);
\draw [rounded corners,->,>=stealth] ($(y_t.west) + (-2.0,0.0)$) node [left] (y_t_1) {$\cdots$} -- (y_t.west) ;
\draw [rounded corners,->,>=stealth] (y_t1.east) -- ($(y_t1.east)+ (0.5,0.0)$) node [right] (y_t2) {$\cdots$};

\draw[decorate,decoration={brace,amplitude=2mm,raise=3mm},yshift=0pt]($(memory_layer_frame.north west) + (0.0,0.1)$) -- ($(memory_layer_frame.north west) + (0.0,1.6)$) node [black,midway,xshift=-1.2cm,yshift=0cm,align=center] {\footnotesize CRF\\ \footnotesize layer};

\end{tikzpicture}
\end{center}
\caption{\label{fig:model}Illustration of \mecrf[s] with a single memory hop, showing the network architecture at time step $t$ and $t+1$.}
\end{figure*}

As illustrated in \figref{fig:model}, \mecrf can be divided into two major parts: (1) the memory layer; and (2) the \crf layer. The memory layer can be further broken down into three main components: (a) the input memory $\vec{m}_{1:t}$; (b) the output memory $\vec{c}_{1:t}$; and (c) the current input $\vec{u}_t$, which represents the current step (also known as the ``question'' in the context of \memnn). The input and output memory representations are connected via an attention mechanism whose weights are determined by measuring the similarity between the input memory and the current input. 
The \crf layer, on the other hand, takes the output of the memory layer as input. In the remainder of this section, we detail the elements of \mecrf.

\subsection{Memory Layer}

\subsubsection{Input memory} 

Every element (word/post) in a sequence $\mathbf{x}$ is encoded with $\vec{x}_{t} = \Phi(x_t)$, where $\Phi(\cdot)$ can be any encoding function mapping the input $x_t$ into a vector $\vec{x}_{t} \in \R^{d}$. This results in the sequence $\{\vec{x}_{1}, \ldots, \vec{x}_{T}\}$.
While this new sequence can be seen as the memory in the context of \memnn[s], one major drawback of this approach, as pointed out by \newcite{Seo+:2016}, is the insensitivity to temporal information between memory cells. We therefore follow \newcite{XiongMS16} in injecting temporal signal into the memory using a bi-directional $\gru$ encoding \cite{Cho+:2014b}: 
\begin{align}
\vecfwd{m}_{t} &= \grufwd(\vec{x}_{t}, \vecfwd{m}_{t-1}) \label{equ:grumfwd} \\ 
\vecrev{m}_{t} &= \grurev(\vec{x}_{t}, \vecrev{m}_{t+1}) \label{equ:grumrev} \\ 
\vec{m}_{t}    &= \tanh(\matfwd[m]{W}\vecfwd{m}_{t} + \matrev[m]{W}\vecrev{m}_{t} + \vec{b}_{m}) \label{equ:grum} 
\end{align}
where $\matfwd[m]{W}$, $\matrev[m]{W}$ and $\vec{b}_{m}$ are learnable parameters. 

\subsubsection{Current input} 

This is used to represent the current step $x_t$, be it a word or a post. As in \memnn[s], we want to enforce the current input to be in the same space as the input memory so that we can determine the attention weight of each element in the memory by measuring the relevance between the two. We denote the current input by $\vec{u}_{t} = \vec{m}_t$.

\subsubsection{Attention} 

To determine the attention value of each element in the memory, we measure the relevance between the current step $\vec{u}_{t}$ and $\vec{m}_{i}$ for $i \in [1, t]$ with a $\softmax$ function:
\begin{equation}
p_{t,i} = \softmax(\vec[\T]{u}_{t}\vec{m}_{i})
\end{equation}
where $\softmax(a_{i}) = \dfrac{e^{a_{i}}}{\sum_{j}e^{a_{j}}}$. 

\subsubsection{Output memory} 

Similar to $\vec{m}_{t}$, $\vec{c}_{t}$ is the output memory, and is calculated analogously but with a different set of parameters in the GRUs and $\tanh$ layers of \equref[s]{equ:grumfwd}, \bracketref{equ:grumrev} and \bracketref{equ:grum}. The output memory is used to generate the final output of the memory layer and fed as input to the \crf layer.

\subsubsection{Memory layer output} 

Once the attention weights have been computed, the memory access controller receives the response $\vec{o}$ in the form of a weighted sum over the output memory representations:
\begin{equation}
\vec{o}_{t} = \sum_{i}p_{t,i}\vec{c}_{i}
\end{equation}
This allows the model to have unrestricted access to elements in previous steps as opposed to a single vector $\vec{h}_{t}$ in RNNs, thereby enabling \mecrf[s] to detect and effectively incorporate long-range dependencies.

\subsubsection{Extension} 

For more challenging tasks requiring complex reasoning capabilities with multiple supporting facts from the memory, the model can be further extended by stacking multiple memory hops, in which case the output of the $k$-th hop is taken as input to the $(k+1)$-th hop:
\begin{equation}
\label{equ:memnn}
\vec[k+1]{u}_{t} = \vec[k]{o}_{t} + \vec[k]{u}_{t}
\end{equation}
where $\vec[k+1]{u}_{t}$ encodes not only information at the current step ($\vec[k]{u}_{t}$) but also relevant knowledge from the memory ($\vec[k]{o}_{t}$). In the scope of this work, we limit the number of hops to 1.

\subsection{\crf Layer}

Once the representation of the current step $\vec[K+1]{u}_t$ is computed --- incorporating relevant information from the memory (assuming the total number of memory hops is $K$) --- it is then fed into a \crf layer:
\begin{equation}
\label{equ:score}
s(\mathbf{x}, \mathbf{y}) = \sum_{t=0}^{T}\mat[y_t,y_{t+1}]{A} + \sum_{t=1}^{T}\mat[t,y_t]{P}
\end{equation}
where $\mat{A} \in \R^{|\mathcal{Y}|\times|\mathcal{Y}|}$ is the \crf transition matrix, $|\mathcal{Y}|$ is the size of the label set, and $\mat{P} \in \R^{T \times |\mathcal{Y}|}$ is a linearly transformed matrix from $\vec[K+1]{u}_{t}$ such that $\mat[t,:]{P}\tran = \mat[s]{W}\vec[K+1]{u}_{t}$ where $\mat[s]{W} \in \R^{|\mathcal{Y}| \times h}$ with $h$ being the size of $\vec{m}_t$. Here, $\mat[i, j]{A}$ represents the score of the transition from the $i$-th tag to the $j$-th tag whereas $\mat[i, j]{P}$ is the score of the $j$-th tag at time $i$. Using the scoring function in \equref{equ:score}, we calculate the score of the sequence $\mathbf{y}$ normalised by the sum of scores of all possible sequences $\tilde{\mathbf{y}}$, and this becomes the probability of the true sequence:
\begin{equation}
p(\mathbf{y}|\mathbf{x}) = \dfrac{\exp(s(\mathbf{x}, \mathbf{y}))}{\sum_{\tilde{\mathbf{y}} \in \mathbf{Y}_{\mathbf{X}}}\exp(s(\mathbf{x}, \tilde{\mathbf{y}}))}
\end{equation}

We train the model to maximise the probability of the gold label sequence with the following loss function:
\begin{equation}
\label{equ:loss}
\mathcal{L} = \sum_{n=1}^{N}\log p(\mathbf{y}^{(n)}|\mathbf{x}^{(n)})
\end{equation}
where $p(\mathbf{y}^{(n)}|\mathbf{x}^{(n)})$ is calculated using the forward--backward algorithm. Note that the model is fully end-to-end differentiable.

At test time, the model predicts the output sequence with maximum a
posteriori probability:
 \begin{equation}
 \mathbf{y}^{*} = \argmax_{\tilde{\mathbf{y}} \in \mathbf{Y}_{\mathbf{x}}}p(\tilde{\mathbf{y}}|\mathbf{x})
 \end{equation}
Since we are only modelling bigram interactions, we adopt the Viterbi algorithm for decoding.


\section{Thread Discourse Structure Prediction}
\label{sec:expforum}

In this section, we describe how \mecrf[s] can be applied to the task of thread discourse structure prediction, wherein we attempt to predict which post(s) a given post directly responds to, and in what way(s) (as captured by dialogue acts). This is a novel approach to this problem and capable of natively handling both tasks within the same network architecture.

%
%

\subsection{Dataset and Task}

In this work, we adopt the dataset of \newcite{kim+:2010},\footnote{\url{http://people.eng.unimelb.edu.au/tbaldwin/resources/conll2010-thread/}} which consists of 315 threads and 1,332 posts, collected from the Operating System, Software, Hardware and Web Development sub-forums of \system{cnet}.\footnote{\url{http://forums.cnet.com/}} Every post has been manually linked to preceding post(s) in the thread that it is a direct response to (in the form of ``links''), and the nature of the response for each link (in the form of ``dialogue acts'', or ``DAs''). In this dataset, it is not uncommon to see messages respond to posts which occur much earlier in the thread (based on the chronological ordering of posts). In fact, 18\% of the posts link to posts other than their immediately preceding post.

The task is defined as follows: given a list of preceding posts ${x_{1}, \ldots, x_{t-1}}$ and the current post $x_{t}$, to classify which posts it links to ($l_{t}$) and the dialogue act ($y_{t}$) of each such link. 
In the scope of this work, \mecrf[s] are capable of modelling both tasks natively, and therefore a natural fit for this problem.

\subsection{Experimental Setup}

In this dataset, in addition to the body of text, each post is also associated with a title. We therefore use two encoders, $\Phi_{\mathit{title}}(\cdot)$ and $\Phi_{\mathit{text}}(\cdot)$, to process them separately and then concatenate  $\vec{x}_t = [\Phi_{\mathit{title}}({x}_{t});\Phi_{\mathit{text}}({x}_{t})]\tran$.
Here, $\Phi_{\mathit{title}}(\cdot)$ and $\Phi_{\mathit{text}}(\cdot)$ take word embeddings as input, processing each post at the word level, as opposed to the post-level bi-directional GRU in \equref[s]{equ:grumfwd} and \bracketref{equ:grumrev}, and the representation of a post $\vec{x}_t$ (either title or text) is obtained by transforming the last and first hidden states of the forward and backward word-level GRU, similar to \equref{equ:grum}. Note that $\Phi_{\mathit{title}}(\cdot)$ and $\Phi_{\mathit{text}}(\cdot)$ do not share parameters. As in \newcite{Tang+:2016}, we further restrict $\vec[k]{m}_i = \vec[k]{c}_i$ to curb overfitting.

In keeping with \newcite{wang2014}, we complement the textual representations with hand-crafted structural features $\Phi_{s}(x_t) \in \R^{2}$:
\begin{compactitem}
  \item initiator: a binary feature indicating whether the author of the current post is the same as the initiator of the thread,
  \item position: the relative position of the current post, as a ratio over the total number of posts in the thread;
\end{compactitem}
Also drawing on \newcite{wang2014}, we incorporate punctuation-based features $\Phi_{p}(x_t) \in \R^{3}$: the number of question marks, exclamation marks and URLs in the current post. The resultant feature vectors are projected into an embedding space by $\mat[s]{W}$ and $\mat[p]{W}$ and concatenated with $\vec{x}_i$, resulting in the new $\vec[\prime]{x}_i$. Subsequently, $\vec[\prime]{x}_i$ is fed into the bi-directional GRUs to obtain $\vec{m}_i$.

For link prediction, we generate a supervision signal from the annotated links, guiding the attention to focus on the correct memory position:
\begin{equation}
\mathcal{L}_{\text{LNK}} = \sum_{n=1}^{N}\sum_{t=1}^{T}\xentropy(\vec[(n)]{l}_t, \vec[(n)]{p}_t)
\end{equation}
where $\vec[(n)]{l}_t$ is a one-hot vector indicating where the link points to for the $t$-th post in the $n$-th thread, and $\vec[(n)]{p}_t = \{p_{t,1},\ldots,p_{t,t}\}$ is the predicted distribution of attention over the $t$ posts in the memory. To accommodate the first post in a thread, as it points to a virtual ``head'' post, we set a dummy post, $\vec{m}_0 = \vec{0}$, of the same size as $\vec{m}_i$. While the dataset contains multi-headed posts (posts with more than one outgoing link), following \newcite{wang2014}, we only include the most recent linked post during training, but evaluate over the full set of labelled links.

For this task, \mecrf is jointly trained to predict both the link and dialogue act with the following loss function:
\begin{equation}
\mathcal{L}^{\prime} = \alpha\mathcal{L}_{\text{DA}} + (1-\alpha)\mathcal{L}_{\text{LNK}}
\end{equation}
where $\mathcal{L}_{\text{DA}}$ is the \crf likelihood defined in \equref{equ:loss}, and $\alpha$ is a hyper-parameter for balancing the emphasis between the two tasks.

Training is carried out with Adam \cite{Kingma+:2015} over 50 epochs with a batch size of 32. We use the following hyper-parameter settings: word embedding size of $20$, $\mat[p]{W} \in \R^{100\times3}$, $\mat[s]{W} \in \R^{50\times2}$, $\alpha=0.5$, hidden size of $\Phi_{\mathit{title}}$ and $\Phi_{\mathit{text}}$ is $20$, hidden size of $\grufwd$ and $\grurev$ is $50$.
Dropout is applied to all $\gru$ recurrent units on the input and output connections with a keep rate of $0.7$.

Lastly, we also explore the idea of curriculum learning \cite{Bengio+:2009}, by fixing the \crf transition matrix $\mat{A} = \mat{0}$ for the first $e=20$ epochs, after which we train the parameters for the remainder of the run. This allows the \mecrf to learn a good strategy for DA and link prediction, as independent ``maxent'' type classifier, before attempting to learn sequence dynamics.  We refer to this variant as ``\mecrf[\textsc{+}]''.

\subsection{Evaluation}

Following \newcite{wang2014}, we evaluate based on post-level micro-averaged F-score. All experiments were carried out with $10$-fold cross validation, stratifying at the thread level. 

We benchmark against the following previous work: the feature-rich CRF-based approach of \newcite{kim+:2010}, where the authors trained independent models for each of link and DA classification (``\kim''); the feature-rich CRF-based approach of \newcite{wang2014}, where the author further extended the feature set of \newcite{kim+:2010} and jointly trained a CRF over the link and DA prediction tasks (``\wangcrf''); and the dependency parser-based approach of \newcite{wang2014}, where the author treated the discourse structure prediction task as a constrained dependency parsing problem, with posts as nodes in the dependency graph, and the constraint that links must connect to preceding posts in the thread (``\wangparser'').\footnote{Note that a mistake was found in the results in the original paper \cite{Wang+:2011}, and we use the corrected results from \newcite{wang2014}.} In addition to the \crf/parser-based systems, we also build a \memnn-based baseline (named \memnn) where \memnn shares the architecture of the memory layer in \mecrf but excludes the use of the \crf layer. Instead, \memnn, following the work of \newcite{Sukhbaatar+:2015}, predicts the final answer by:
\begin{equation}
\label{equ:finalpred}
\hat{\vec{y}} = \softmax(\mat[\text{DA}]{W}(\vec[K+1]{u}))
\end{equation}
where $\hat{\vec{y}}$ is the predicted DA distribution, $\mat[\text{DA}]{W} \in \R^{|\mathcal{Y}| \times d}$ is a parameter matrix for the model to learn, and $K=1$ is the total number of hops. This is equivalent to classifying link and DA independently at each time step $t$ without taking transitions between DA labels into account.

\subsection{Results}

The experimental results are presented in \tabref{tbl:performance}, wherein the first three rows are the three baseline systems. 

\begin{table}[tb]
	\centering
	\begin{tabular}{lccc}
		\toprule
		Model & Link & DA & Joint \\
		\midrule
		\kim & 86.3 & 75.1 & --- \\
		\wangcrf & 82.3 & 73.4 & 66.5 \\
		\wangparser & 85.0 & 75.7 & 70.6 \\ \addlinespace[1ex]
		\memnn & 85.8 & 76.0 & 69.5 \\ \addlinespace[1ex]
		\mecrf & \textbf{86.4} & \textbf{77.5} & 70.9 \\
		\mecrf[\textsc{+}] & 86.3 & 77.4 & \textbf{71.2} \\
		\bottomrule
	\end{tabular}
	\caption{Post-level Link and DA F-scores. Performance for \mecrf and \mecrf[\textsc{+}] is marco-averaged over 5 runs.}
\label{tbl:performance}
\end{table}

\paragraph{State-of-the-art post-level results.} \mecrf[s] achieve state of the art results in terms of joint post-level F-score, substantially better than the baselines. 
While \mecrf slightly outperforms the current state-of-the-art (\wangparser), \mecrf[\textsc{+}] improves the performance and achieves a further 0.3\% absolute gain.

\paragraph{Curriculum learning improves joint prediction.} Despite the slight performance drop on the DA and link prediction tasks, \mecrf[\textsc{+}], with the \crf transition matrix frozen for the first $20$ epochs, achieves a $\sim$0.3\% absolute gain in joint F-score over \mecrf.
This suggests that the sequence dynamics between posts, while difficult to capture, are beneficial to the overall task (resulting in more coherent DA and link predictions) if trained with proper initialisation.

\paragraph{\memnn vs.\ \mecrf[s].} We see consistent gains across all three tasks when the CRF layer is added. Although not presented in \tabref{tbl:performance}, the difference is most notable at the thread-level (i.e.\ a thread is correct iff all posts are tagged correctly), highlighting the importance of sequential transitional information between posts.

\paragraph{\crf vs.\ \mecrf[s].} Note that \kim is not trained jointly on the two component tasks, but individually on each task. Without additional data, jointly training on the two tasks generates results that are comparable or substantially better over the individual tasks. This highlights the effectiveness of \mecrf, especially with the link prediction performance comparable to that of a single-task model \crf, and surpassing it in the case of DA.

\subsection{Analysis}

\begin{figure}[tb]
\begin{tikzpicture}
    \begin{axis}[
        width  = \columnwidth,
        height = 6cm,
        major x tick style = transparent,
        ybar=2*\pgflinewidth,
        bar width=7pt,
        ymajorgrids = true,
        ylabel = {Link\&DA Joint F-score},
        xlabel = {Post depth},
        symbolic x coords={1,3,5,7,9},
        xtick = data,
        title = Link\&DA Joint,
        xticklabels={{[1,2]},{[3,4]},{[5,6]},{[7,8]},{[9,)}},
        scaled y ticks = false,
        enlarge x limits=0.15,
        ymin=0,
        legend cell align=left,
        legend style={
                font=\fontsize{7.5}{2}\selectfont,
                at={(0.98,0.98)},
                anchor=north east,
                column sep=1ex
        }
    ]
    \addplot[style={black,fill=white,mark=none}]
         coordinates {(1,94.7) (3,50.7) (5,40.3) (7,28.5) (9,27.6)};
    \addplot[style={black,fill=black!40,mark=none}]
         coordinates {(1,95.3) (3,60.1) (5,49.7) (7,36.9) (9,34.2)};
    \addplot[style={black,fill=black,mark=none}]
         coordinates {(1,86.4) (3,53.4) (5,40.5) (7,43.2) (9,43.7)};
    \legend{\wangcrf,\wangparser,\mecrf[\textsc{+}]}
    \end{axis}
\end{tikzpicture}

\caption{Breakdown of post-level Joint F-scores by post depth, where e.g.\ ``$[1,2]$'' is the joint F-score over posts of depth 1--2, \ie the first or second post in the thread. Note that we take the reported performance of \wangcrf and \wangparser from \newcite{wang2014}.}
\label{fig:jointbreakdown}
\end{figure}
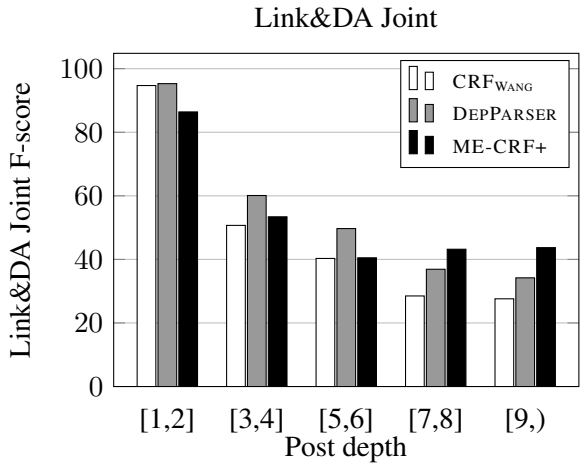

\begin{figure}[tb]
\begin{tikzpicture}
    \begin{axis}[
        width  = \columnwidth,
        height = 4cm,
        major x tick style = transparent,
        ybar=2*\pgflinewidth,
        bar width=7pt,
        ymajorgrids = true,
        ylabel = {Link F-score},
        xlabel = {Post depth},
        symbolic x coords={1,3,5,7,9},
        xtick = data,
        title = Link,
        xticklabels={{[1,2]},{[3,4]},{[5,6]},{[7,8]},{[9,)}},
        scaled y ticks = false,
        enlarge x limits=0.15,
        ymin=40,
        ymax=108,
        legend columns=-1,
        legend cell align=left,
        legend style={
                font=\fontsize{7.5}{2}\selectfont,
                at={(0.5,1.35)},
                anchor=south,
                column sep=1ex
        }
    ]
    \addplot[style={black,fill=white,mark=none}]
         coordinates {(1,100) (3,74.0) (5,64.6) (7,53.6) (9,55.0)};
    \addplot[style={black,fill=black!40,mark=none}]
         coordinates {(1,100) (3,78.8) (5,72.9) (7,63.3) (9,53.6)};
    \addplot[style={black,fill=black,mark=none}]
         coordinates {(1,94.2) (3,77.2) (5,71.4) (7,64.7) (9,63.9)};
    \legend{\system{CRFSGD},\system{MaltParser},\mecrf[\textsc{+}]};
    \end{axis}
\end{tikzpicture}

\begin{tikzpicture}
    \begin{axis}[
        width  = \columnwidth,
        height = 4cm,
        major x tick style = transparent,
        ybar=2*\pgflinewidth,
        bar width=7pt,
        ymajorgrids = true,
        ylabel = {DA F-score},
        xlabel = {Post depth},
        title = DA,
        symbolic x coords={1,3,5,7,9},
        xtick = data,
        xticklabels={{[1,2]},{[3,4]},{[5,6]},{[7,8]},{[9,)}},
        scaled y ticks = false,
        enlarge x limits=0.15,
        ymin=40,
        ymax=108,
        legend columns=-1,
        legend cell align=left,
        legend style={
                font=\fontsize{7.5}{2}\selectfont,
                at={(0.5,1.05)},
                anchor=south,
                column sep=1ex
        }
    ]
    \addplot[style={black,fill=white,mark=none}]
         coordinates {(1,94.6) (3,60.9) (5,55.0) (7,42.9) (9,47.7)};
    \addplot[style={black,fill=black!40,mark=none}]
         coordinates {(1,95.7) (3,62.2) (5,59.8) (7,49.6) (9,53.6)};
    \addplot[style={black,fill=black,mark=none}]
         coordinates {(1,89.1) (3,61.8) (5,53.9) (7,55.1) (9,61.0)};
    \end{axis}
\end{tikzpicture}

\caption{\label{fig:breakdown}Breakdown of post-level Link and DA F-scores by post depth.}
\end{figure}
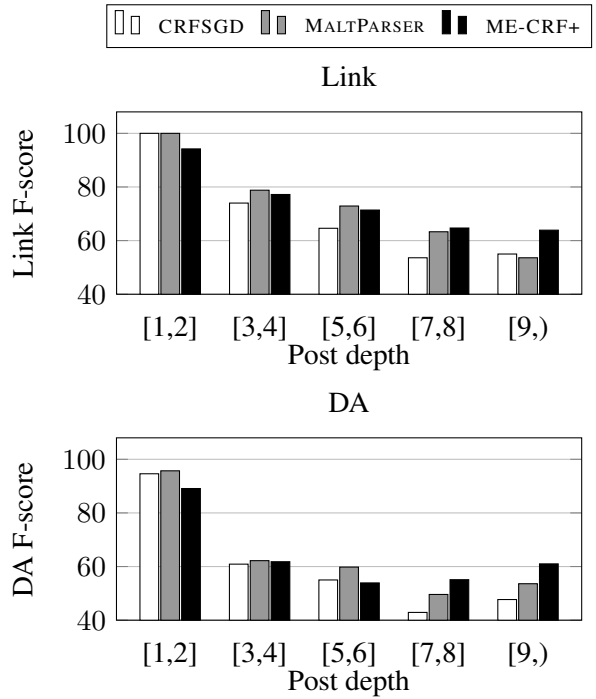

We break the performance down by the depth of each post in a given thread, and present the results in \figref{fig:jointbreakdown}. Although below the baselines for the interval $[1,2]$, \mecrf[\textsc{+}] consistently outperforms \wangcrf from depth $3$ onwards, and is superior to \wangparser for depths $[7,)$. Breaking down the performance further to the individual tasks of Link and DA prediction, as displayed in \figref{fig:breakdown}, we observe a similar trend. In line with the findings in the work of \newcite{wang2014}, this confirms that prediction becomes progressively more difficult as threads grow longer, which is largely due to the increased variability in discourse structure. Despite the escalated difficulty, \mecrf[\textsc{+}] is substantially superior to the baselines when classifying deeper posts. 


Between the \crf[-based] models, it is worth noting that despite the lower performance for $[1,2]$, \mecrf[\textsc{+}] benefits from having global access to the entire sequence, and consistently outperforms \wangcrf for depths $[3,)$, highlighting the effectiveness of the memory mechanism.
Overall, these results validate our hypothesis that having unrestricted access to the whole sequence is beneficial, especially for long-range dependencies, offering further evidence of the power of \mecrf[s].





\section{Named Entity Recognition}
\label{sec:expner}

In this section, we present experiments in a second setting: named entity recognition, over the CoNLL 2003 English NER shared task dataset \cite{Tjong+:2003}. Our interest here is in evaluating the ability of \mecrf to capture document context, to aid in the identification and disambiguation of NEs.

\subsection{Dataset and Task}

The CoNLL 2003 NER shared task dataset consists of $14,041/3,250/3,453$ sentences in the training/development/test set, resp., extracted from $946/216/231$ Reuters news articles from the period 1996--97. The goal is to identify individual token occurrences of NEs, and tag each with its class (\eg \class{LOCATION} or \class{ORGANISATION}). Here, we use the IOB tagging scheme.
In terms of tagging schemes, while some have shown improvements with a more expressive \system{IOBES} marginally \cite{Ratinov+:2009,Dai+:2015}, we stick to the \system{BIO} scheme for simplicity and the observation of little improvement between these schemes by \newcite{Lample+:2016}.

\subsection{Experimental Setup}

We choose $\Phi(x_t)$ to be a lookup function, returning the corresponding embedding $\vec{x}_t$ of the word $x_t$. In addition to the word features, we employ a subset of the lexical features described in \newcite{Huang+:2015}, based on whether the word:
\begin{itemize}[noitemsep,topsep=0pt]
	\item starts with a capital letter;
	\item is composed of all capital letters;
	\item is composed of all lower case letters;
	\item contains non initial capital letters;
	\item contains both letters and digits;
	\item contains punctuation.
\end{itemize}
These features are all binary and refered to as $\Phi_{l}(x_t)$. Similar to the thread structure prediction experiments, we concatenate $\Phi_{l}(x_t)$ with $\vec{x}_{t}$ to generate the new input $\vec[\prime]{x}$ to the bi-directional GRUs in \equref[s]{equ:grumfwd} and \bracketref{equ:grumrev}.

In order to incorporate information in the document beyond sentence boundaries, we encode every word sequentially in a document with $\Phi$ and $\grufwd$ and $\grurev$, and store them in the memory $\vec{m}_i$ and $\vec{c}_i$ for $i \in [1,t^\prime]$, where $t^\prime$ is the index of the current word $t$ in the document.

Training is carried out with Adam, over $100$ epochs with a batch size of 32. We use the following hyper-parameter settings: word embedding size = $50$; hidden size of $\grufwd$ and $\grurev$ = $50$; $\matfwd[m]{W}$ and $\matrev[m]{W} \in \R^{50\times50}$; and $\vec{b}_{m} \in \R^{50}$. Dropout is applied to all $\gru$ recurrent units on the input and output connections, with a keep rate of $0.8$. We initialise \mecrf with pre-trained word embeddings and keep them fixed during training. While we report results only on the test set, we use early stopping based on the development set.

\subsection{Evaluation}

Evaluation is based on span-level NE F-score, based on the official
CoNLL evaluation script.\footnote{\url{http://www.cnts.ua.ac.be/conll2000/chunking/conlleval.txt}}

We compare against the following baselines:
\begin{compactenum}
\item a \crf over hand-tuned lexical features (``\crf'': \newcite{Huang+:2015})
\item an \lstm and bi-directional \lstm (``\lstm'' and ``\bilstm'', resp.: \newcite{Huang+:2015})
\item a \crf taking features from a convolutional neural network as
  input (``\convcrf'': \newcite{Collobert+:2011})
\item a \crf over the output of either a simple \lstm or bidirectional \lstm (``\lstmcrf'' and ``\bilstmcrf'', resp.: \newcite{Huang+:2015})
\end{compactenum}
Note that for our word embeddings, while we observe better performance
with \system{GloVe} \cite{Pennington+:2014}, for fair comparison
purposes, we adopt the same \system{Senna} embeddings
\cite{Collobert+:2011} as are used in the baseline
methods.\footnote{\newcite{Lample+:2016} report a higher result of  90.9 using a \bilstmcrf architecture, but augmented with skip $n$-grams \cite{Ling+:2015a} and character embeddings. Due to the differing underlying representation, we exclude it from the comparison.}

\subsection{Results}

The experimental results are presented in \tabref{tbl:nerperformance}. Results for the baseline methods are based on the published results of \newcite{Huang+:2015} and \newcite{Collobert+:2011}.
Note that none of the systems in \tabref{tbl:nerperformance} use external gazetteers, to make the comparison fair. 
As can be observed, \mecrf achieves the best performance, beating all the baselines. 

\begin{table}[tb]
	\centering
	\begin{tabular}{lc}
		\toprule
		Model & F-score \\
		\midrule
		\crf & 86.1 \\
		\lstm & 83.7 \\
		\bilstm & 85.2 \\ 
		\convcrf & 88.7 \\ \addlinespace[1ex]
		\lstmcrf & 88.4 \\
		\bilstmcrf & 88.8 \\ \addlinespace[1ex]
		\mecrf & \textbf{89.5} \\
		\bottomrule
	\end{tabular}
	\caption{NER performance on the CoNLL 2003 English NER shared task dataset.}
        \label{tbl:nerperformance}
\end{table}


To gain a better understanding of what the model has learned, \tabref{tbl:nerexample} presents two examples where \mecrf focuses on words beyond the current sentence boundaries. In the example on the left, where the target word is \ex{Juventus} (an Italian soccer team), \mecrf directs the attention mainly to the occurrence of the same word in a previous sentence and a small fraction to \ex{Manchester} (a UK soccer team, in this context). Note that it does not attend to the other NE (\ex{Europe}) in that sentence, which is of a different NE class. In the example on the right, on the other hand, \mecrf allocates attention to the same words as the target word in the current sentence. Note that the second occurrence of \ex{Interfax} in the memory is the same occurrence as the first word in the current sentence. While more weight is placed on the second \ex{Interfax}, close to one third of the attention is also asigned to the first occurrence. Given that the memory, $\vec{m}_i$ and $\vec{c}_i$, is encoded with bi-directional GRUs, the first \ex{Interfax} should, to some degree, capture the succeeding neighbouring elements: \ex{news agency}. 

This is reminiscent of label consistency in the works of \newcite{Sutton+:2004} and \newcite{Finkel+:2005}, but differs in that the consistency constraint is soft as opposed to hard in previous studies, and automatically learned without the use of heuristics. 

\begin{table}[tb]
\center
\resizebox{.9\linewidth}{!}{
\begin{minipage}{.40\columnwidth}
\begin{tabular}{lc}
\toprule
Memory & $p_{t,i}$\\
\midrule
$\ldots$ & \\
Manchester & \cellcolor{red!23} 0.23 \\
United & 0.00 \\
face & 0.00 \\
Juventus & \cellcolor{red!65} 0.65 \\
in & 0.00 \\
Europe & 0.00 \\
$\ldots$ & \\
\midrule
\multicolumn{2}{l}{European champions}\\
\multicolumn{2}{l}{\underline{Juventus} $\ldots$}\\
\bottomrule
\end{tabular}
\end{minipage}%
\quad\quad\quad
\begin{minipage}{.40\columnwidth}
\begin{tabular}{lc}
\toprule
Memory & $p_{t,i}$\\
\midrule
$\ldots$ & \\
, & 0.00 \\
Interfax & \cellcolor{red!32} 0.32 \\
news & 0.00 \\
agency & 0.00 \\
said & 0.00 \\
. & 0.00 \\\hdashline
\underline{Interfax} & \cellcolor{red!68} 0.68 \\
\midrule
\multicolumn{2}{l}{\underline{Interfax} quoted}\\
\multicolumn{2}{l}{Russian $\ldots$}\\
\bottomrule
\end{tabular}
\end{minipage}
}
\caption{An NER example showing learned attention to long-range contextual dependencies. The last row is the current sentence. The underlined words indicate the target word $x_{t}$, and the dashed line indicates a sentence boundary.}
\label{tbl:nerexample}
\end{table}


\section{Conclusion}
\label{sec:conc}

In this paper, we have presented \mecrf, a model extending linear-chain \crf[s] by including external memory. 
This allows the model to look beyond neighbouring items and access long-range context.
Experimental results demonstrate the effectiveness of the proposed method over two tasks: forum thread discourse analysis, and named entity recognition.


\section*{Acknowledgments}

We thank the anonymous reviewers for their valuable feedback, and gratefully acknowledge the support of Australian Government Research Training Program Scholarship and National Computational Infrastructure (NCI Australia). This work was also supported in part by the Australian Research Council.

\bibliography{ijcnlp2017}

\begin{thebibliography}{}
\expandafter\ifx\csname natexlab\endcsname\relax\def\natexlab#1{#1}\fi

\bibitem[{Bahdanau et~al.(2014)Bahdanau, Cho, and Bengio}]{Bahdanau+:2014}
Dzmitry Bahdanau, Kyunghyun Cho, and Yoshua Bengio. 2014.
\newblock Neural machine translation by jointly learning to align and
  translate.
\newblock In {\em Proceedings of the 3rd International Conference on Learning
  Representations (ICLR 2015)\/}. San Diego, USA.

\bibitem[{Bengio et~al.(2009)Bengio, Louradour, Collobert, and
  Weston}]{Bengio+:2009}
Yoshua Bengio, J{\'e}r\^{o}me Louradour, Ronan Collobert, and Jason Weston.
  2009.
\newblock Curriculum learning.
\newblock In {\em Proceedings of the 26th Annual International Conference on
  Machine Learning (ICML 2009)\/}. Montreal, Canada, pages 41--48.

\bibitem[{Bengio et~al.(1994)Bengio, Simard, and Frasconi}]{Bengio+:1994}
Yoshua Bengio, Patrice Simard, and Paolo Frasconi. 1994.
\newblock Learning long-term dependencies with gradient descent is difficult.
\newblock {\em IEEE Transactions on Neural Networks\/} 5(2):157--166.

\bibitem[{Cho et~al.(2014)Cho, Van~Merri{\"e}nboer, Bahdanau, and
  Bengio}]{Cho+:2014b}
Kyunghyun Cho, Bart Van~Merri{\"e}nboer, Dzmitry Bahdanau, and Yoshua Bengio.
  2014.
\newblock On the properties of neural machine translation: Encoder-decoder
  approaches.
\newblock In {\em Proceedings of SSST-8, Eighth Workshop on Syntax, Semantics
  and Structure in Statistical Translation\/}. Doha, Qatar, pages 103--111.

\bibitem[{Collobert et~al.(2011)Collobert, Weston, Bottou, Karlen, Kavukcuoglu,
  and Kuksa}]{Collobert+:2011}
Ronan Collobert, Jason Weston, L{\'e}on Bottou, Michael Karlen, Koray
  Kavukcuoglu, and Pavel Kuksa. 2011.
\newblock Natural language processing (almost) from scratch.
\newblock {\em Journal of Machine Learning Research\/} 12:2493--2537.

\bibitem[{Dai et~al.(2015)Dai, Lai, Chang, and Tsai}]{Dai+:2015}
Hong-Jie Dai, Po-Ting Lai, Yung-Chun Chang, and Richard Tzong-Han Tsai. 2015.
\newblock Enhancing of chemical compound and drug name recognition using
  representative tag scheme and fine-grained tokenization.
\newblock {\em Journal of Cheminformatics\/} 7(1):S14.

\bibitem[{Finkel et~al.(2005)Finkel, Grenager, and Manning}]{Finkel+:2005}
Jenny~Rose Finkel, Trond Grenager, and Christopher Manning. 2005.
\newblock Incorporating non-local information into information extraction
  systems by gibbs sampling.
\newblock In {\em Proceedings of the 43rd Annual Meeting of the Association for
  Computational Linguistics (ACL 2005)\/}. Ann Arbor, USA, pages 363--370.

\bibitem[{Geman and Geman(1984)}]{Geman+:1984}
Stuart Geman and Donald Geman. 1984.
\newblock Stochastic relaxation, {Gibbs} distributions, and the {Bayesian}
  restoration of images.
\newblock {\em IEEE Transactions on Pattern Analysis and Machine
  Intelligence\/} 6:721--741.

\bibitem[{Graves et~al.(2013)Graves, Mohamed, and Hinton}]{Graves+:2013}
Alex Graves, Abdel-rahman Mohamed, and Geoffrey Hinton. 2013.
\newblock Speech recognition with deep recurrent neural networks.
\newblock In {\em 2013 IEEE International Conference on Acoustics, Speech and
  Signal Processing (ICASSP 2013)\/}. Vancouver, Canada, pages 6645--6649.

\bibitem[{Hochreiter et~al.(2001)Hochreiter, Bengio, Frasconi, and
  Schmidhuber}]{Hochreiter+:2001}
Sepp Hochreiter, Yoshua Bengio, Paolo Frasconi, and J{\"u}rgen Schmidhuber.
  2001.
\newblock Gradient flow in recurrent nets: the difficulty of learning long-term
  dependencies.

\bibitem[{Hochreiter and Schmidhuber(1997)}]{Hochreiter+:1997}
Sepp Hochreiter and J{\"u}rgen Schmidhuber. 1997.
\newblock Long short-term memory.
\newblock {\em Neural computation\/} 9(8):1735--1780.

\bibitem[{Huang et~al.(2015)Huang, Xu, and Yu}]{Huang+:2015}
Zhiheng Huang, Wei Xu, and Kai Yu. 2015.
\newblock Bidirectional lstm-crf models for sequence tagging.
\newblock {\em arXiv preprint arXiv:1508.01991\/} .

\bibitem[{Kim et~al.(2010)Kim, Wang, and Baldwin}]{kim+:2010}
Su~Nam Kim, Li~Wang, and Timothy Baldwin. 2010.
\newblock Tagging and linking web forum posts.
\newblock In {\em Proceedings of the Fourteenth Conference on Computational
  Natural Language Learning (CoNLL 2010)\/}. Association for Computational
  Linguistics, Uppsala, Sweden, pages 192--202.

\bibitem[{Kingma and Ba(2015)}]{Kingma+:2015}
Diederik Kingma and Jimmy Ba. 2015.
\newblock Adam: A method for stochastic optimization.
\newblock In {\em Proceedings of the 3th International Conference on Learning
  Representations (ICLR 2015)\/}. San Diego, USA.

\bibitem[{Krishnan and Manning(2006)}]{Krishnan+:2006}
Vijay Krishnan and Christopher~D. Manning. 2006.
\newblock An effective two-stage model for exploiting non-local dependencies in
  named entity recognition.
\newblock In {\em Proceedings of the 21st International Conference on
  Computational Linguistics and 44th Annual Meeting of the Association for
  Computational Linguistics (ACL 2006)\/}. Sydney, Australia, pages 1121--1128.

\bibitem[{Lafferty et~al.(2001)Lafferty, McCallum, and
  Pereira}]{Lafferty+:2001}
John~D. Lafferty, Andrew McCallum, and Fernando C.~N. Pereira. 2001.
\newblock Conditional random fields: Probabilistic models for segmenting and
  labeling sequence data.
\newblock In {\em Proceedings of the Eighteenth International Conference on
  Machine Learning (ICML 2001)\/}. San Francisco, USA, pages 282--289.

\bibitem[{Lai et~al.(2015)Lai, Xu, Liu, and Zhao}]{Lai+:2015}
Siwei Lai, Liheng Xu, Kang Liu, and Jun Zhao. 2015.
\newblock Recurrent convolutional neural networks for text classification.
\newblock In {\em Proceedings of the 29th AAAI Conference on Artificial
  Intelligence (AAAI 2015)\/}. Austin, USA, volume 333, pages 2267--2273.

\bibitem[{Lample et~al.(2016)Lample, Ballesteros, Subramanian, Kawakami, and
  Dyer}]{Lample+:2016}
Guillaume Lample, Miguel Ballesteros, Sandeep Subramanian, Kazuya Kawakami, and
  Chris Dyer. 2016.
\newblock Neural architectures for named entity recognition.
\newblock In {\em Proceedings of the 2016 Conference of the North American
  Chapter of the Association for Computational Linguistics: Human Language
  Technologies (NAACL 2016)\/}. San Diego, USA, pages 260--270.

\bibitem[{Liao and Grishman(2010)}]{Liao+:2010}
Shasha Liao and Ralph Grishman. 2010.
\newblock Using document level cross-event inference to improve event
  extraction.
\newblock In {\em Proceedings of the 48th Annual Meeting of the Association for
  Computational Linguistics (ACL 2010)\/}. Uppsala, Sweden, pages 789--797.

\bibitem[{Ling et~al.(2015)Ling, Tsvetkov, Amir, Fermandez, Dyer, Black,
  Trancoso, and Lin}]{Ling+:2015a}
Wang Ling, Yulia Tsvetkov, Silvio Amir, Ramon Fermandez, Chris Dyer, Alan~W
  Black, Isabel Trancoso, and Chu-Cheng Lin. 2015.
\newblock Not all contexts are created equal: Better word representations with
  variable attention.
\newblock In {\em Proceedings of the 2015 Conference on Empirical Methods in
  Natural Language Processing (EMNLP 2015)\/}. Lisbon, Portugal, pages
  1367--1372.

\bibitem[{Linzen et~al.(2016)Linzen, Dupoux, and Goldberg}]{Linzhen+:2016}
Tal Linzen, Emmanuel Dupoux, and Yoav Goldberg. 2016.
\newblock Assessing the ability of lstms to learn syntax-sensitive
  dependencies.
\newblock {\em Transactions of the Association for Computational Linguistics
  (TACL 2016)\/} 4:521--535.

\bibitem[{Pennington et~al.(2014)Pennington, Socher, and
  Manning}]{Pennington+:2014}
Jeffrey Pennington, Richard Socher, and Christopher Manning. 2014.
\newblock {GloVe}: Global vectors for word representation.
\newblock In {\em Proceedings of the 2014 Conference on Empirical Methods in
  Natural Language Processing (EMNLP 2014)\/}. Doha, Qatar, pages 1532--1543.

\bibitem[{Ratinov and Roth(2009)}]{Ratinov+:2009}
Lev Ratinov and Dan Roth. 2009.
\newblock Design challenges and misconceptions in named entity recognition.
\newblock In {\em Proceedings of the Thirteenth Conference on Computational
  Natural Language Learning (CoNLL 2009)\/}. Boulder, USA, pages 147--155.

\bibitem[{Seo et~al.(2017)Seo, Min, Farhadi, and Hajishirzi}]{Seo+:2016}
Minjoon Seo, Sewon Min, Ali Farhadi, and Hannaneh Hajishirzi. 2017.
\newblock Query-reduction networks for question answering.
\newblock In {\em Proceedings of the 5th International Conference on Learning
  Representations (ICLR 2017)\/}. Toulon, France.

\bibitem[{Sukhbaatar et~al.(2015)Sukhbaatar, Szlam, Weston, and
  Fergus}]{Sukhbaatar+:2015}
Sainbayar Sukhbaatar, Arthur Szlam, Jason Weston, and Rob Fergus. 2015.
\newblock End-to-end memory networks.
\newblock In {\em Proceedings of Advances in Neural Information Processing
  Systems (NIPS 2015)\/}. Montr\'{e}al, Canada, pages 2440--2448.

\bibitem[{Sutton and McCallum(2004)}]{Sutton+:2004}
Charles Sutton and Andrew McCallum. 2004.
\newblock Collective segmentation and labeling of distant entities in
  information extraction.
\newblock Technical report, Massachusetts Univ Amherst Dept of Computer
  Science.

\bibitem[{Tang et~al.(2016)Tang, Qin, and Liu}]{Tang+:2016}
Duyu Tang, Bing Qin, and Ting Liu. 2016.
\newblock Aspect level sentiment classification with deep memory network.
\newblock In {\em Proceedings of the 2016 Conference on Empirical Methods in
  Natural Language Processing (EMNLP 2016)\/}. Austin, USA, pages 214--224.

\bibitem[{Tjong Kim~Sang and De~Meulder(2003)}]{Tjong+:2003}
Erik~F Tjong Kim~Sang and Fien De~Meulder. 2003.
\newblock Introduction to the conll-2003 shared task: Language-independent
  named entity recognition.
\newblock In {\em Proceedings of the seventh conference on North American
  Chapter of the Association for Computational Linguistics (NAACL 2003)\/}.
  Edmonton, Canada, pages 142--147.

\bibitem[{Wang(2014)}]{wang2014}
Li~Wang. 2014.
\newblock {\em Knowledge discovery and extraction of domain-specific web
  data\/}.
\newblock Ph.D. thesis, The University of Melbourne.

\bibitem[{Wang et~al.(2011)Wang, Lui, Kim, Nivre, and Baldwin}]{Wang+:2011}
Li~Wang, Marco Lui, Su~Nam Kim, Joakim Nivre, and Timothy Baldwin. 2011.
\newblock Predicting thread discourse structure over technical web forums.
\newblock In {\em Proceedings of the 2011 Conference on Empirical Methods in
  Natural Language Processing (EMNLP 2011)\/}. Edinburgh, UK, pages 13--25.

\bibitem[{Weston et~al.(2015)Weston, Chopra, and Bordes}]{Weston+:2015}
Jason Weston, Sumit Chopra, and Antoine Bordes. 2015.
\newblock Memory networks.
\newblock In {\em Proceedings of the 3rd International Conference on Learning
  Representations (ICLR 2015)\/}. San Diego, USA.

\bibitem[{Xiong et~al.(2016)Xiong, Merity, and Socher}]{XiongMS16}
Caiming Xiong, Stephen Merity, and Richard Socher. 2016.
\newblock Dynamic memory networks for visual and textual question answering.
\newblock In {\em Proceedings of the 33rd International Conference on Machine
  Learning (ICML 2016)\/}. New York, USA, pages 2397--2406.

\bibitem[{Yang et~al.(2016)Yang, Yang, Dyer, He, Smola, and Hovy}]{Yang+:2016}
Zichao Yang, Diyi Yang, Chris Dyer, Xiaodong He, Alex Smola, and Eduard Hovy.
  2016.
\newblock Hierarchical attention networks for document classification.
\newblock In {\em Proceedings of the 2016 Conference of the North American
  Chapter of the Association for Computational Linguistics --- Human Language
  Technologies (NAACL HLT 2016)\/}. San Diego, USA, pages 1480--1489.

\bibitem[{Zhang et~al.(2017)Zhang, Culbertson, and Paritosh}]{Zhang+:2017}
Amy Zhang, Bryan Culbertson, and Praveen Paritosh. 2017.
\newblock Characterizing online discussion using coarse discourse sequences.
\newblock In {\em Proceedings of the 11th AAAI International Conference on Web
  and Social Media (ICWSM 2017)\/}. Palo Alto, USA, pages 357--366.

\end{thebibliography}
\bibliographystyle{ijcnlp2017}

\end{document}